\documentclass[conference]{IEEEtran}
\IEEEoverridecommandlockouts
\usepackage{amsmath,amssymb,amsfonts}
\usepackage{algorithmic}
\usepackage{graphicx}
\usepackage{textcomp}
\usepackage{xcolor}
\usepackage{times}  
\usepackage{helvet}  
\usepackage{courier}  
\usepackage[hyphens]{url}  
\usepackage{graphicx} 
\usepackage{amsmath,amssymb,amsfonts}
\usepackage{algorithmic}
\usepackage{graphicx}
\usepackage{textcomp}
\usepackage{colortbl}
 \usepackage{multirow}
\usepackage{amsthm}
\usepackage{diagbox} 
\usepackage{colortbl} 
\usepackage{diagbox} 
\usepackage{makecell} 
\usepackage{tabularx}
\usepackage{cite}
\usepackage{booktabs} 
\usepackage{multirow}
\usepackage{caption}
\newcolumntype{C}[1]{>{\centering\arraybackslash}p{#1}}
\def\BibTeX{{\rm B\kern-.05em{\sc i\kern-.025em b}\kern-.08em
    T\kern-.1667em\lower.7ex\hbox{E}\kern-.125emX}}
\begin{document}

\title{HyperGCL: Multi-Modal Graph Contrastive Learning via Learnable Hypergraph Views
}
\author{
    \IEEEauthorblockN{Khaled Mohammed Saifuddin\IEEEauthorrefmark{1},  Shihao Ji\IEEEauthorrefmark{2}, Esra Akbas\IEEEauthorrefmark{3} }  
    \IEEEauthorblockA{\IEEEauthorrefmark{1} Northeastern University, Boston, MA, USA \\
                      \IEEEauthorrefmark{2} University of Connecticut, Storrs, CT, USA \\
                      \IEEEauthorrefmark{3} Georgia State University, Atlanta, GA, USA }  
    \IEEEauthorblockA{\IEEEauthorrefmark{1} k.saifuddin@northeastern.edu, 
                      \IEEEauthorrefmark{3} shihao.ji@uconn.edu,  \IEEEauthorrefmark{2} eakbas1@gsu.edu}  
}

\newcommand{\shygan}{\texttt{SHyGAN}}
\newcommand{\hcl}{\texttt{HCL}}
\newcommand{\hygan}{\texttt{HyGAN}}
\newcommand{\hypergcl}{\texttt{HyperGCL}}
\newcommand{\netcl}{\texttt{NetCL}}
\newcommand{\negset}{\texttt{NegS}}
\newcommand{\posset}{\texttt{PosS}}
\newcommand{\hypergcls}{\texttt{HyperGCL}$_{sim}$}
\newcommand{\hypergcld}{\texttt{HyperGCL}$_{dis}$}
\newcommand{\hypergclal}{\texttt{HyperGCL}$_{al}$}
\newcommand{\hypergclag}{\texttt{HyperGCL}$_{ag}$}
\newcommand{\hypergcllg}{\texttt{HyperGCL}$_{lg}$}
\newcommand{\hypergclwaug}{\texttt{HyperGCL}$_{w/o Aug}$}
\newcommand{\hypergclwnetcl}{\texttt{HyperGCL}$_{w/o NetCL}$}
\maketitle

\begin{abstract}
Recent advancements in Graph Contrastive Learning (GCL) have demonstrated remarkable effectiveness in improving graph representations. However, relying on predefined augmentations (e.g., node dropping, edge perturbation, attribute masking) may result in the loss of task-relevant information and a lack of adaptability to diverse input data. Furthermore, the selection of negative samples remains rarely explored.
In this paper, we introduce \hypergcl, a novel multimodal GCL framework from a hypergraph perspective. \hypergcl\ constructs three distinct hypergraph views by jointly utilizing the input graph’s structure and attributes, enabling a comprehensive integration of multiple modalities in contrastive learning. A learnable adaptive topology augmentation technique enhances these views by preserving important relations and filtering out noise. View-specific encoders capture essential characteristics from each view, while a network-aware contrastive loss leverages the underlying topology to define positive and negative samples effectively. 
Extensive experiments on benchmark datasets demonstrate that \hypergcl\ achieves state-of-the-art node classification performance.

\end{abstract}
%

\section{Introduction} \label{intro}

Building on the success of contrastive learning (CL) in computer vision and natural language processing \cite{tian2020contrastive, gao2021simcse}, CL approaches have been extended to graph data—known as Graph Contrastive Learning (GCL)—where Graph Neural Networks (GNNs) learn robust representations by maximizing agreement between augmented graph views \cite{velickovic2018deep, peng2020graph, hassani2020contrastive, suresh2021adversarial}. Nonetheless, current GCL methods still exhibit several limitations.

\textbf{First}, they often depend on handcrafted augmentations such as node dropping, edge perturbation, and attribute masking. While these techniques can be effective, they risk discarding crucial task-relevant information and force models to rely on specific hyperparameter settings~\cite{You2020GraphCL, thakoor2021large, zeng2021contrastive}. \textbf{Second}, most methods primarily treat the graph structure and node attributes as a single, unified source of information \cite{velickovic2018deep, peng2020graph}, overlooking the distinct yet complementary roles that topology and attribute can play in uncovering complex patterns. 
\textbf{Third}, they generally focus on local pairwise (dyadic) relationships, limiting their ability to capture higher-order global patterns \cite{peng2020graph, zhu2021graph, shen2023neighbor}.
\textbf{Fourth}, many GCL strategies employ contrastive losses originally designed for image data, often overlooking the distinct characteristics of graph-structured data, such as the homophily principle \cite{zhu2020beyond}. These approaches typically treat all non-positive nodes as negative examples, resulting in a large number of negative samples and leading to high computational and memory overhead~\cite{zhu2021graph, You2020GraphCL}.

To address these issues, we propose \hypergcl, a multimodal attribute and structure-aware GCL framework from a hypergraph perspective. Hypergraphs naturally model complex systems and can capture hidden higher-order information present in networks even in standard graphs.
Unlike prior GCL models that often treat graph structure and its attributes as a unified source of information and mainly focus on dyadic relations, our approach considers them as two distinct modalities and generates different hypergraph views for CL.
Specifically, we design three distinct hypergraph views from graph structure and its attributes to capture different granularities of higher-order information for CL. These views include an attribute-driven hypergraph that leverages existing attributes of nodes representing semantic information and two structure-driven hypergraphs (local and global) that leverage varying graph structural information. This multimodal design allows \hypergcl\ to effectively capture various perspectives from the input graph, providing a richer and more robust framework for representation learning.
Moreover, instead of using predefined augmentation techniques, we utilize an adaptive model for each hypergraph view using a learnable Gumbel-Softmax function \cite{jang2016categorical}. This introduces controlled stochasticity, enhancing the diversity and quality of training samples for CL. The dynamic adjustment of the augmentation process improves the discriminative power of the views by selectively highlighting key relationships within the hypergraph.

We employ view-specific encoders for each augmented view. For attribute-driven hypergraph view, we use the Hypergraph Attention Network (\hygan) \cite{ding2020more, hwang2021hyfer} that learns node embeddings by identifying semantically important nodes and hyperedges. However, since \hygan\ focuses on semantic features, directly applying it to structure-driven hypergraphs may result in the loss of structural information. To address this, we introduce Structure-aware \hygan\ (\shygan), a specialized variant that incorporates node structure information in the input layer and structural biases in the attention layers, ensuring the capture of both semantic and structural information.

Unlike traditional GCL methods that adopt contrastive losses tailored for computer vision, such as InfoNCE \cite{oord2018representation} or NT-Xent \cite{chen2020simple}, we introduce a novel network-aware contrastive loss, \netcl. This loss builds upon NT-Xent by leveraging the network topology to define positive and negative samples more effectively. 
To further optimize the selection of negative samples, we propose two selective negative sampling strategies:
(1) \textit{distance-based}, and (2) \textit{similarity-based}.
These strategies reduce the size of the negative sample while maintaining its effectiveness in the contrastive framework. 

The contributions of this work are summarized as follows:
\begin{itemize}

\item \textbf{Multimodal Hypergraph View Generation and Adaptive Augmentation:} In  \hypergcl, we introduce a novel multimodal framework that generates three hypergraph views to capture diverse granularities of structural and attribute-driven information. To ensure robustness, we propose a learnable augmentation mechanism using a Gumbel-Softmax function, eliminating the need for predefined augmentations.

\item \textbf{View-Specific Encoder:} \hypergcl\ employs view-specific encoders to learn view embeddings. We apply \hygan\ for attribute-driven hypergraphs and introduce \shygan\ for structure-driven hypergraphs to capture both semantic and structural information effectively.

\item \textbf{Network-Aware Contrastive Loss (\netcl):} We propose \netcl, a topology-guided contrastive loss that leverages network structure to define positive and negative samples, aligning multimodal hypergraph representations while optimizing computational efficiency.

\end{itemize}

\begin{figure*}[t!]
\includegraphics[width=18.5cm, height=8cm]{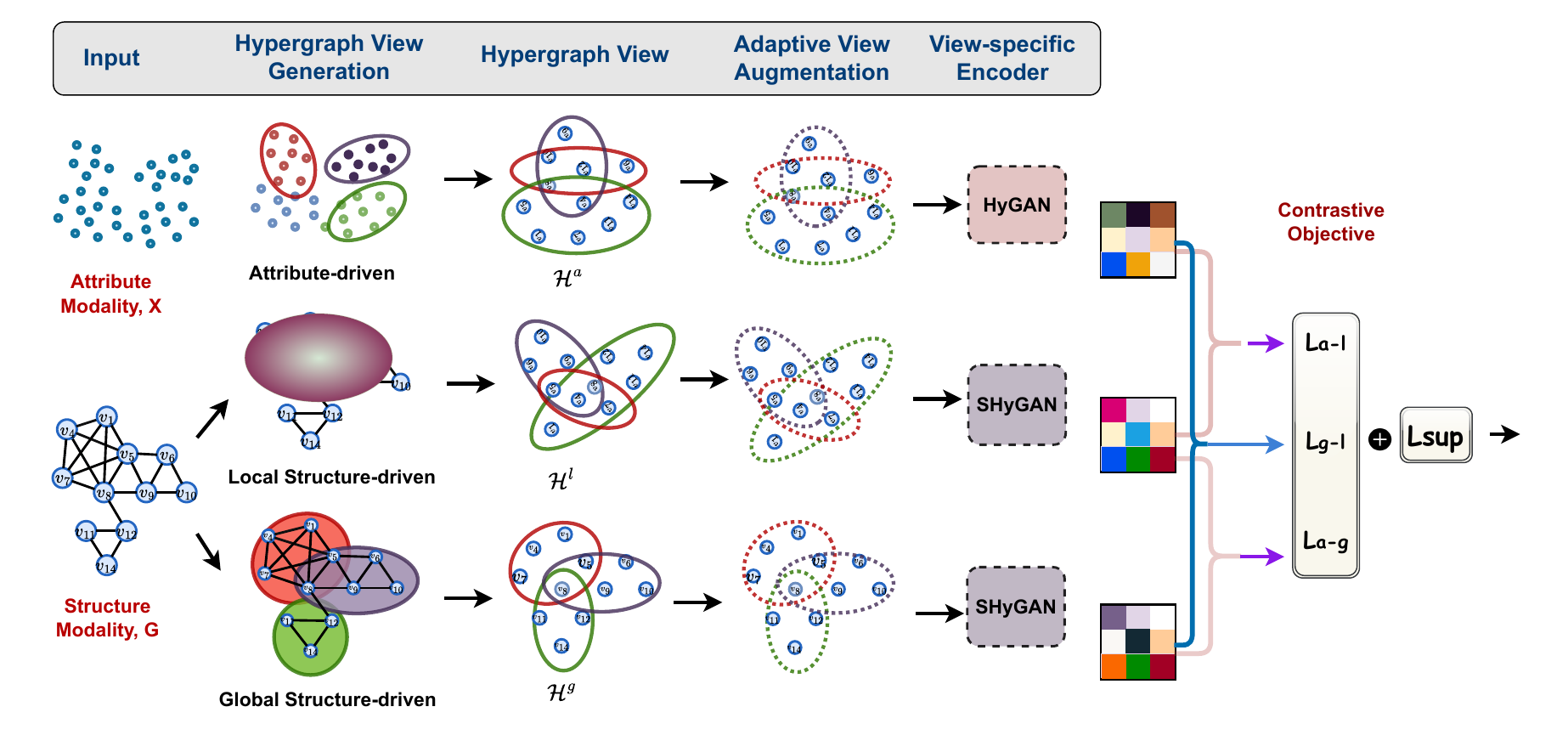}\vspace{-2mm}
   \caption{System architecture of \hypergcl. After constructing three different hypergraph views from the input graph and node attributes, we exploit a learnable view augmentation technique to generate adaptive views. View-specific encoders are used to learn each view and finally, a network-aware contrastive loss is used with a supervised loss to train the model. }
    \label{fig:method} \vspace{-4mm}
\end{figure*}

\vspace{-2mm}
\section{Related Work}

\textbf{Contrastive Learning} (CL) has become a powerful paradigm for graph representation learning by maximizing agreement across augmented views~\cite{Zhu:2020vf}. Early work such as DGI~\cite{velivckovic2018deep}, inspired by Deep InfoMax~\cite{hjelm2018learning}, maximizes mutual information between local patches and a global summary. Later approaches generate multiple views using various augmentations, e.g., feature/edge masking~\cite{thakoor2021large}, node dropping, subgraph extraction~\cite{You2020GraphCL}, node/edge insertion/deletion~\cite{zeng2021contrastive}, or graph diffusion~\cite{hassani2020contrastive,ma2023self}. However, most GCL methods still employ contrastive loss functions from computer vision, typically treating the same node in different views as a positive sample and all other nodes as negatives, thereby overlooking inherent graph topology~\cite{zhu2021graph,You2020GraphCL}.

\textbf{Hypergraph Neural Networks} (HyperGNNs) extend GNNs to model complex relationships through hyperedges connecting multiple nodes. HGNNs \cite{feng2019hypergraph} and HyperGCN \cite{yadati2019hypergcn} were pioneers, applying spectral convolution to hypergraphs using clique expansion and hypergraph Laplacians.
Moreover, attention-based models like HAN \cite{chen2020hypergraph} and HyperGAT \cite{ding2020more} adaptively learn node and hyperedge importance. HyperSAGE \cite{arya2020hypersage} and UniGNN \cite{huang2021unignn} avoid information loss by directly performing message passing on hypergraphs. The AllSetTransformer \cite{chien2021you} combines Deep Sets \cite{zaheer2017deep} and Set Transformers \cite{lee2019set} for enhanced flexibility and expressive power.

\section{Methodology}
This section outlines the components of \hypergcl: Hypergraph View generation and adaptive augmentation, View-Specific Hypergraph Encoder, and Network-Aware Contrastive Loss (NetCL). Figure \ref{fig:method} presents the system architecture. 

\subsection{Multimodal Hypergraph View Generation and Adaptive Augmentation}

Graphs are effective for modeling pairwise relationships between nodes but often fail to capture complex higher-order interactions. Moreover, most graph learning methods treat the graph structure and the node attributes as a unified source of information. This blending can obscure the distinct yet complementary roles that topology and attributes play in uncovering intricate patterns. To address this, we treat the graph structure \( G = (V, E) \)—where \( V \) denotes the set of nodes and \( E \) represents the edges— and node attributes \( \mathbf{X} \in \mathbb{R}^{|V| \times d} \), where \( d \) is the feature dimension as distinct modalities. By leveraging these two modalities, we generate multiple hypergraph views for contrastive learning: attribute-driven, local structure-driven, and global structure-driven hypergraphs. Each view captures unique granularities, uncovering higher-order interactions while preserving and integrating the multimodal information inherent in topology and attributes for richer representations.

\subsubsection{Attribute-driven Hypergraph View (\(\mathcal{H}^{a}\))} \label{attribute_driven_hyG}
To incorporate attribute information, we construct an \emph{attribute-driven hypergraph} by grouping semantically similar nodes into hyperedges. Specifically, we apply both \(k\)-nearest neighbors (\(k\)-NN) and \(k\)-means clustering to the node attributes \(\mathbf{X}\). Using \(k\)-NN, each node \(v_i \in V\) and its \(k\) closest neighbors form an initial hyperedge. Formally, for each node \( v_i \), a hyperedge \( \bar{e_j} \) is formed as:
\vspace{-1mm}
\[
\bar{e_{j}} = \{v_i\} \cup \{v_k \in V \mid v_k \text{ is } \text{a } k \text{ nearest neighbor of } v_i\}.\vspace{-1mm}
\] 

Additionally, the clusters derived from \(k\)-means are also used to create hyperedges. Let \(\mathcal{C} = \{C_1, C_2, \ldots, C_k\}\) be the set of clusters obtained from \(k\)-means. Each cluster \(C_c\) is treated as a hyperedge \(\bar{\bar{e}}_c\), where \(\bar{\bar{e}}_c = C_c\). Each node \(v_i\) is then assigned to its \(s\) nearest clusters, chosen based on the smallest Euclidean distances from \(v_i\) to the respective cluster centers. Formally, for each node \(v_i\), the set of hyperedges it belongs to is given by:
\vspace{-1mm}
\[
\begin{split}
e_{j}^{a} = \{\bar{e_{j}} \mid  v_i \in \bar{e_{j}}\} \cup \{\bar{\bar{e_{j}}} \mid v_i  \in C_j \text{ and } \\
C_j \text{ is one of the } s \text{ closest clusters to } v_i\}.
\end{split}\vspace{-1mm}
\]

Combining all hyperedges produced by \(k\)-NN and \(k\)-means yields the attribute-driven hypergraph \(\mathcal{H}^{a} = (V, \mathcal{E}^{a})\), where \(\mathcal{E}^{a}\) is the set of all constructed hyperedges. This framework effectively captures higher-order relationships among nodes based on their semantic similarities.

\subsubsection{Structure-driven Hypergraph Views}
While attribute-driven hypergraphs effectively preserve node semantic similarity, they often overlook the original graph’s structural properties. To address this gap, we construct two additional hypergraph views---\emph{local} and \emph{global}---that capture different levels of structural information.

\textit{i. Local Structure-driven Hypergraph View (\(\mathcal{H}^l\)).}
To capture local structural context, we represent each node’s 1-hop ego network as a hyperedge. Specifically, for each node \(v_i \in V\), we form the hyperedge \(e_j^l\) by including \(v_i\) and all of its 1-hop neighbors. Formally,
\vspace{-1mm}
\[
e_j^l = \{v_i\} \cup \{v_k \in V \mid (v_i, v_k) \in E\}.
\]

Combining all such hyperedges yields the hypergraph \(\mathcal{H}^l = (V, \mathcal{E}^l)\), where \(\mathcal{E}^l\) is the set of hyperedges derived from each node’s ego network. This construction preserves the local context of each node within the hypergraph, allowing for a nuanced analysis of connectivity patterns.
\\
\textit{ii. Global Structure-driven Hypergraph View (\(\mathcal{H}^g\)).}
To capture higher-order structural relationships on a global scale, we form hyperedges from subgraphs that extend beyond local neighborhoods. Although there are multiple ways to define such higher-order structures (e.g., cliques, motifs), we focus on communities---densely interconnected subgraphs that capture group-level interactions. Concretely, let 
\(\mathcal{CM} = \{CM_1, CM_2, \ldots, CM_k\}\)
denote a set of communities. Each community \(CM_c\) is then represented as a hyperedge by its constituent nodes as: $
\bar{\bar{\bar{e}}}_{c} = CM_c.$
By collecting all such hyperedges, we obtain the global structure-driven hypergraph \(\mathcal{H}^g = (V, \mathcal{E}^g)\), where \(\mathcal{E}^g\) is the set of hyperedges derived from the identified communities. This construction encapsulates broad structural patterns in the graph, complementing the local perspective captured by \(\mathcal{H}^l\).

When representing each community as a hyperedge, it is crucial to ensure overlaps between communities to maintain the connectivity of the hypergraph. Therefore, we explore various overlapping community detection methods and ultimately adopt the algorithm described in \cite{chen2010detecting}, as it demonstrated superior performance in our experiments. This algorithm uses edge attributes as weights. In cases where the input graph lacks these attributes, we assign a uniform weight of 1 to each edge. By gathering all identified communities as hyperedges, we form the hypergraph \( \mathcal{H}^g = (V, \mathcal{E}^g) \), where \( \mathcal{E}^g \) represents the set of all hyperedges derived from communities.

It is important to note that even after applying the overlapping community detection algorithm, some communities may remain isolated, resulting in hyperedges that are not interconnected. This isolation can stop the flow of information within the hypergraph. To address this and enhance connectivity, we incorporate global nodes selected based on their high closeness centrality scores from the input graph. Closeness centrality, which measures the average shortest distance of a node to all other nodes, helps identify nodes that can efficiently disseminate information across the network. By linking these centrally located global nodes to hyperedges, we significantly improve the interconnectivity and information exchange across the hypergraph, ensuring a more cohesive and efficient network structure.


\subsubsection{Adaptive View Augmentation} \label{adaptive_view_augmentation}

In each hypergraph view, we introduce a learnable Gumbel-Softmax function to adaptively augment the hyperedges. We begin by initializing learnable logits that represent the probability of each node's association with a particular hyperedge. For each node \(v_i\), we perturb its logits \(\mathbf{\phi}_{v_i}\) with Gumbel noise \(\mathbf{\epsilon}\) and apply the Softmax function as follows:
$\mathbf{p}_{v_i} = \text{softmax}\left( \frac{\mathbf{\phi}_{v_i} + \mathbf{\epsilon}}{\tau} \right)$,
where \( \tau \) is the temperature parameter. The binary mask for node \(v_i\) is obtained by applying a threshold \(\theta\) to \( \mathbf{p}_{v_i} \) as 
\(\mathbf{m}_{v_i} = \left(\mathbf{p}_{v_i} > \theta\right).
\)
To ensure gradients propagate through the Softmax probabilities \( \mathbf{p}_{v_i} \) instead of the binary mask \( \mathbf{m}_{v_i} \), we employ the straight-through estimator as \(
\mathbf{\tilde{m}}_{v_i} = (\mathbf{m}_{v_i} - \mathbf{p}_{v_i}) + \mathbf{p}_{v_i}.
\)
The final augmented hypergraph view is produced by element-wise multiplying the binary mask matrix \( \mathbf{M} \), composed of all \( \mathbf{m} \), with the original hypergraph incidence matrix \( \mathbf{A} \) as 
$\mathbf{\tilde{A}} = \mathbf{M} \odot \mathbf{A}.$ 
This adaptive augmentation technique also works as a method for refining the hypergraph views. 
By leveraging this learnable Gumbel-Softmax-based augmentation strategy, our approach ensures the generation of diverse samples, enhancing the effectiveness of CL in hypergraph settings.

\subsection{View-Specific Hypergraph Encoder} \label{hyG_encoder}
To generate node embeddings from each view, we utilize view-specific encoders capturing different granularities of information. We apply \hygan\ to the attribute-driven hypergraph view generating node embedding matrix $\mathbf{Z}^a$. Similarly, we define \shygan\ to apply on the local structure-driven hypergraph view and the global structure-driven hypergraph view, generating the node embedding matrices $\mathbf{Z}^l$ and $\mathbf{Z}^g$, respectively. These embeddings are then utilized in contrastive loss functions to preserve different granularities of information.
\\
\textbf{\hygan}: Motivated by \cite{ding2020more,hwang2021hyfer}, we employ the Hypergraph Attention Network (\hygan) on the attribute-driven hypergraph view focusing on capturing attribute-based semantic information. \hygan\ accomplish this by employing a two-level attention mechanism: \textit{node-to-hyperedge level attention} and \textit{hyperedge-to-node level attention}.

While the \textit{node-to-hyperedge level attention} mechanism aggregates node information into hyperedge representations, 
it also pinpoints which nodes carry higher semantic importance within each hyperedge and assigns them greater weight during aggregation. 
Concretely, the representation of hyperedge \(e_j\) at the \(l\)-th layer, denoted by \(q_{e_j}^l\), is formulated as:
\begin{equation}
\begin{aligned}
\label{eqn_conditional_q}
q_{e_j}^{l} &=  \sum_{v_{i} \in e_{j}} \Big[\Gamma_{ji} \, W_{1} \, p_{v_i}^{l-1} \Big],
\end{aligned}
\end{equation}
\begin{equation}
\label{eq_Gamma_ji}
\text{where, }\Gamma_{ji} = \frac{\exp(r_{ji})}{\sum_{v_{k} \in e_j} \exp(r_{jk})}, 
\end{equation}

\begin{equation}
\label{eq_r_ij_l}
r_{ji}^l = \frac{1}{\sqrt{d_{\mathrm{hid}}}} \, \beta\Bigl(\mathbf{W}_{2} \, p_{v_i}^{l-1} \;\odot\; \mathbf{W}_{3} \, q_{e_j}^{l-1}\Bigr).
\end{equation}
Here, \(r_{ji}^{l}\) is the attention coefficient of node \(v_i\) in hyperedge \(e_j\) at the \(l\)-th layer. The function \(\beta\) is a non-linear activation, each \(\mathbf{W}\) represents a trainable weight matrix, 
\(p_{v_i}\) and \(q_{e_j}\) are the representations of node \(v_i\) and hyperedge \(e_j\), $d_{hid}$ is their hidden dimension, and \(\odot\) is the Hadamard product.

\noindent

Similarly, the \textit{hyperedge-to-node level attention} mechanism aggregates hyperedges to generate node representations. 
It employs an attention mechanism to identify hyperedges that are semantically important for each node and assign them more weights during aggregation. 
Formally, the representation of node \(v_i\) at the \(l\)-th layer, denoted \(p_{v_i}^{l}\), is defined as:
\begin{equation}
\begin{aligned}
p_{v_i}^{l} 
&=  \sum_{e_{j} \in E_{i}} \Big[\Lambda_{ij} \, W_{4} \, q_{e_j}^{l} \Big],
\end{aligned}
\end{equation}

\begin{equation}
\label{eq_Lambda_ij}
\text{where, }\Lambda_{ij} = \frac{\exp(y_{ij})}{\sum_{e_{k} \in E_{i}} \exp(y_{ik})},
\end{equation}

\begin{equation}
\label{eq_y_ij_l}
\text{and, }y_{ij}^l = \frac{1}{\sqrt{d_{\mathrm{hid}}}} \, \beta\Bigl(\mathbf{W}_{5} \, q_{e_j}^{l} \,\odot\, \mathbf{W}_{6} \, p_{v_i}^{l-1}\Bigr).
\end{equation}
Here, \(y_{ij}^l\) denotes the attention coefficient of hyperedge \(e_j\) on node \(v_i\) at the \(l\)-th layer. 
\\
\textbf{Structure-aware \hygan\ (\shygan)}: While \hygan\ focuses on attribute-based semantic features to identify important nodes and hyperedges, this approach can lead to a loss of structural information when applied directly to the structure-driven hypergraph. To address this limitation, we introduce a specialized variant of \hygan\ called Structure-aware \hygan\ (\shygan) by introducing a two-level topology-guided attention network. \shygan\ leverages structural inductive biases in the attention layers to identify significant nodes and hyperedges from both semantic and structural perspectives. Additionally,  \shygan\ incorporates learnable nodes' structural feature encoding to enhance the initial node features. 

\subsubsection{Node's Structural Feature Encoding} A node's significance in graph data is defined by its connectivity and role within the graph's structure, not just its individual attributes, where regular models often miss these distinctions. 
To capture these structural details, we introduce three structure encoding techniques: (1) Local Connectivity Encoding ($lce$), (2) Centrality Encoding ($ce$), and (3) Distinctiveness Encoding ($de$). These are combined with the initial node features \( x_{v_i}^0 \) of each node \( v_i \) to enrich the overall representation as follows:
\begin{equation}
    \begin{aligned}
        x_{v_i} = \text{Sum}(x_{v_i}^0, lce_{v_i}, ce_{v_i}, de_{v_i}).
    \end{aligned}
\end{equation}
\textit{i. Local Connectivity Encoding: }When we represent each community as a hyperedge, we risk losing important local connectivity information between the nodes, which may be vital for accurate hyperedge representation. To address this issue, we apply a Graph Convolutional Network (GCN) to the input graph, capturing crucial local connectivity patterns. Specifically, the local connectivity encoding for each node \(v_i\) is computed as: $lce_{v_i} = \mathcal{G}{\text{conn}}\left( v_i, \mathbb{N}(v_i); \Phi \right)$, where $lce_{v_i}$ represents the local connectivity encoding for node $v_i$, derived using $\mathcal{G}_{\text{conn}}$, a GCN function, which processes the neighborhood $\mathbb{N}(v_i)$ with trainable parameters $\Phi$.
\\
\textit{ii. Centrality Encoding:} To capture the role and influence of each node, we incorporate \emph{closeness centrality} into the node features. Nodes with higher closeness centrality are closer to all other nodes in the graph, indicating that they can disseminate information more efficiently. We map these centrality scores into embedding vectors via a learnable function \(\mathcal{G}_{\text{central}}\), defined as: $ce = \mathcal{G}_{\text{central}}(c; \psi)$,
where \( c \) is the vector of nodes' centrality scores, and \( \psi \) is a learnable parameter.
\\
\textit{iii. Distinctiveness Encoding:}
Nodes appearing in multiple hyperedges may lose distinctiveness, reducing their significance. We define a distinctiveness score \({d}\) for each node \(v_i\) as:
${d}_{v_i} = 1 - \left(\frac{|{\mathcal{E}_{v_i}}|}{|{\mathcal{E}}|}\right)$,
where $|\mathcal{E}_{v_i}|$ is the number of hyperedges node \(v_i\) belongs to, and $|\mathcal{E}|$ is the total number of hyperedges. Higher counts result in lower Distinctiveness scores. We generate an distinctiveness encoding \(de\) for each node, via a learnable function \( \mathcal{G}_{\text{Distinct}}\) defined as
$de = \mathcal{G}_{\text{Distinct}}(d; \zeta)$,
where \(d\) is the vector of nodes' Distinctiveness scores, and \(\zeta\) is a learnable parameter.

\subsubsection{Topology-Guided Attention} We design topology-guided attention that employs structural inductive biases in the attention layers, enabling the model to identify key nodes and hyperedges from both semantic and structural perspectives. We define two structural inductive biases.
\\
{\textit{i. Node Importance via Local Clustering Coefficient.}}
As a measure of node importance, the local clustering coefficient (\(lc\)) for a given node \(v_i\) within a community \(e_j\) quantifies the density of connections among its neighbors. It is defined as the ratio of the actual number of connections among the neighbors, \(I_{ji}\), to the maximum possible connections within the community. Mathematically, it is expressed as:
\[
lc_{ji} = \frac{2I_{ji}}{g_{ji}(g_{ji} - 1)},
\]
where \(g_{ji}\) represents the degree of node \(v_i\) in the subgraph associated with hyperedge \(e_j\). This metric captures the extent of tightly-knit clusters around a node, reflecting its role in facilitating enhanced information flow within the network.

We incorporate 
$lc$ as a structural inductive bias into Equation \ref{eq_r_ij_l} as follows:
\begin{equation}
\begin{aligned}
r_{ji} 
&= \frac{1}{\sqrt{d_{\mathrm{hid}}}} \, \beta\Bigl(\mathbf{W}_{2}\, p_{v_i}^{l-1} \;\odot\; \mathbf{W}_{3}\, q_{e_j}^{l-1}\Bigr) + lc_{ji}.
\end{aligned}
\end{equation}
\textit{{ii. Hyperedge Importance via Density Score}}
The structural significance of hyperedges can be quantified by evaluating their connectivity and cohesion within a hypergraph. Hyperedges containing more nodes are generally regarded as more influential for a given node compared to those with fewer nodes to which it belongs. We formalize this by defining \emph{hyperedge density}, \(hd\), which measures the fraction of the number of nodes \( m_{e_j} \)  within a hyperedge \( e_j \) relative to the total number of nodes \( m \) in the hypergraph. Formally represented as $hd = \frac{m_{e_j}}{m}.$
A higher \( hd \) value signifies greater interconnectivity among the nodes within the hyperedge, indicating a more cohesive and significant group. 
We integrate \( hd \) as a structural inductive bias into Equation \ref{eq_y_ij_l} as follows:
\begin{equation}
\begin{aligned}
y_{ij} 
&= \frac{1}{\sqrt{d_{\mathrm{hid}}}} \, \beta\Bigl(\mathbf{W}_{5}\, q_{e_j}^{l} \;\odot\; \mathbf{W}_{6}\, p_{v_i}^{l-1}\Bigr) + hd_{ij}.
\end{aligned}
\end{equation}

\subsection{Network-Aware Contrastive Loss (\netcl)}\label{network_aware_cl}
We propose a novel network-aware contrastive loss, termed \netcl, via 
incorporating network topology as supervised signals to define positive and negative samples in \hypergcl. Specifically, instead of forming only a single positive pair per anchor as in regular CL models, \netcl\ allows for multiple positives per anchor. These multiple positives are defined as follows:

{Positive Samples} (\posset) for a node \( v_i \) include the same node \( v_i \) in two different views, nodes that are neighbors of \( v_i \) within the input graph, and nodes that belong to the same hyperedges as \( v_i \) in at least one of the views. Formally, they can be defined as:
\[
\begin{alignedat}{2}
\text{\posset}_{v_i} = & \{ \text{same node in two different views} \} \\
& \cup \{ v_j \mid v_j \text{ is a neighbor of } v_i \text{ in the input graph} \} \\
& \cup \{ v_k \mid v_k \text{ belongs to the same hyperedges as } v_i  \\
& \phantom{\cup \{} \text{in one of the views} \}\}.
\end{alignedat}
\]

Conversely, {Negative Samples} (\negset) for a node \( v_i \) include all other nodes that do not meet these criteria, defined as \(\text{\negset}_{v_i}\).  Considering all these \(\negset\) instances is computationally expensive. To address this, we propose \textit{Distance-based} and \textit{Similarity-based } negative sampling strategies. In the \textit{Distance-based} negative sampling strategy, for an anchor node \(v_i\), we select the top \( t \) nodes from \(\negset_{v_i}\) that are the farthest node from the anchor in the input graph. 
The set of distance-based negative samples for anchor node \(v_i\) is denoted as \(\mathcal{N}_{dis}(v_i)\).
In the \textit{Similarity-based} negative sampling strategy, the top \( t \) nodes from \(\negset_{v_i}\) are selected based on having the lowest cosine similarity to the anchor node \(v_i\), ensuring they are the least semantically similar.
The set of similarity-based negative samples  for the anchor node \(v_i\) is denoted as \(\mathcal{N}_{sim}(v_i)\).
The contrastive loss can then be applied using negative samples selected via either of these strategies. This approach provides a computationally efficient and comprehensive framework.

\begin{table}[b!]
\centering \vspace{-5mm}
\caption{Dataset Statistics. \#N, \#E, and \#C represent the number of nodes, edges, and classes, respectively. Additionally, \#$\mathcal{E}^a$, \#$\mathcal{E}^l$, and \#$\mathcal{E}^g$ denote the number of hyperedges in  $\mathcal{H}^a$, $\mathcal{H}^l$, and $\mathcal{H}^g$.}

\renewcommand{\arraystretch}{1.2}
\setlength{\tabcolsep}{3pt}
\begin{tabular}{|c|c|c|c||c|c|c|}
    \hline
    \cellcolor{gray!60} \textbf{Dataset} & \cellcolor{gray!60} \textbf{\#N} & \cellcolor{gray!60} \textbf{\#E} & \cellcolor{gray!60} \textbf{\#C} & \cellcolor{gray!60} \textbf{\#$\mathcal{E}^a$} & \cellcolor{gray!60} \textbf{\#$\mathcal{E}^l$} & \cellcolor{gray!60} \textbf{\#$\mathcal{E}^g$} \\
    \hline
    \cellcolor{gray!60} \textbf{Cora} & 2,708 & 5,429 & 7 & 2758 & 2708 & 263 \\
    \hline
    \cellcolor{gray!60} \textbf{CS} & 3,312 & 4,715 & 6 & 3362 & 3312 & 563 \\
    \hline
    \cellcolor{gray!60} \textbf{Wiki} & 2,405 & 17,981 & 17 & 2455 & 2405 & 59 \\
    \hline
    \cellcolor{gray!60} \textbf{PT} & 1,912 & 64,510 & 2 & 1962 & 1912 & 112 \\
    \hline
    \cellcolor{gray!60} \textbf{LFMA} & 7,624 & 55,612 & 18 & 7674 & 7624 & 46 \\
    \hline
\end{tabular}
\label{dataset_statistics}
\vspace{-4mm}
\end{table}
\begin{table*}[t!]
\addtolength{\tabcolsep}{1pt}
\centering
\caption{Performance Comparisons: Mean accuracy (\%) \text{$\pm$} standard deviation}
\normalsize
\begin{tabular}{|c|c|c|c|c|c|c|c|c|}
\hline
\cellcolor{gray!60}\textbf{Method}&
\cellcolor{gray!60} \textbf{Model} &
\cellcolor{gray!60} \textbf{Cora} &
\cellcolor{gray!60} \textbf{CS} &
\cellcolor{gray!60} \textbf{Wiki} &
\cellcolor{gray!60} \textbf{PT} &
\cellcolor{gray!60} \textbf{LFMA}
\\
\hline
\multirow{5}{*}{Graph-based}& GCN & 80.88$\pm$1.23 &67.65$\pm$0.72 & 60.66$\pm$1.82 & 65.85$\pm$1.40 & 80.23$\pm$1.08\\
& GAT & 81.08$\pm$0.30 &68.32$\pm$0.80 &61.79$\pm$0.78& 66.30$\pm$0.25 & 82.21$\pm$0.75\\
& GraphSage & 80.64$\pm$0.39 &69.28$\pm$0.66 & 60.17$\pm$0.88& 63.35$\pm$1.22 & 79.66$\pm$1.45\\
& DGI & 81.70$\pm$1.60 &71.50$\pm$0.70 & 64.89$\pm$1.17& 66.82$\pm$1.05 & 83.17$\pm$0.33\\
& GMI & 82.70$\pm$1.20 &73.0$\pm$1.30 & 66.12$\pm$0.65& 66.98$\pm$0.83 & 83.55$\pm$1.74\\
& MVGRL & 82.90$\pm$0.70 &72.60$\pm$1.70 & 66.78$\pm$1.15& 67.18$\pm$0.46 & 84.65$\pm$0.41\\
& GraphCL & 82.50$\pm$1.20 & 72.80$\pm$0.30 & 67.32$\pm$0.66 & 67.58$\pm$0.64 & 83.28$\pm$0.60 \\
& GraphMAE & 83.80$\pm$0.40 &72.40$\pm$0.40 &67.93$\pm$ 0.75& 67.92 $\pm$0.71 & 84.01$\pm$0.57\\
\hline
\multirow{10}{*}{Hypergraph-based} & HGNN & 71.31$\pm$1.66& 65.12$\pm$1.73 & 65.24$\pm$1.10 &66.41 $\pm$0.75 & 78.26$\pm$1.21\\
&HCHA & 71.41$\pm$1.32 & 65.43$\pm$1.15 & 64.41$\pm$ 0.62 & 63.52$\pm$0.80 & 79.44$\pm$1.27\\
& HyperGCN & 60.96$\pm$1.49 & 53.20$\pm$1.53 & 65.84$\pm$ 0.67 & 62.44$\pm$0.68 & 77.89$\pm$1.28\\
& DHGNN & 72.22$\pm$0.92 & 64.59$\pm$1.32 & 65.87$\pm$ 0.86& 65.37$\pm$1.06& 77.22$\pm$0.80\\

& HNHN & 65.76$\pm$0.99 & 63.93$\pm$1.12 & 63.92$\pm$ 1.30 &66.12$\pm$1.26 & 81.17$\pm$1.30\\
& UniGCNII & 70.20$\pm$1.37 &65.57$\pm$1.11 & 66.25$\pm$ 1.15 & 64.24$\pm$0.86 &80.49$\pm$1.54\\
& AllSetTransformer & 70.99$\pm$1.72 &66.60$\pm$1.38 & 67.44$\pm$ 0.88& 65.15$\pm$1.05 & 82.42$\pm$0.95\\
 & DHKH & 64.21$\pm$1.17 &66.34$\pm$1.31 &66.50$\pm$0.80 &  67.04$\pm$0.82&  80.25$\pm$1.25\\
& {\hypergcls} & {84.38}$\pm${0.68} & {71.35}$\pm${0.72} & {68.11}$\pm${0.66} & 
{68.88}$\pm${0.86}& 
{84.12}$\pm${0.42}\\

& \textbf{\hypergcld} & \textbf{85.88}$\pm$\textbf{0.30} & \textbf{73.12}$\pm$\textbf{0.56} & \textbf{69.22}$\pm$\textbf{0.44} & 
\textbf{70.10}$\pm$\textbf{0.25}& 
\textbf{85.15}$\pm$\textbf{1.12}\\
\hline
\end{tabular}
\label{result} \vspace{-4mm}
\end{table*} 
In this paper, to capture and preserve various granularities of information within the node embeddings produced by the encoders, we employ three distinct contrastive learning modules, which are
i) Contrast between the attribute-driven view and the local structure-driven view, ii) Contrast between the global structure-driven view and the attribute-driven view, iii) Contrast between the local structure-driven view and the global structure-driven view.

After obtaining the node embeddings \( \mathbf{Z}^a \) and \( \mathbf{Z}^l \) from attribute-driven and local structure-driven hypergraphs, respectively, we adopt InfoNCE \cite{gutmann2010noise} to estimate the lower bound of the mutual information between them. By defining positive and negative samples, the contrastive loss function can be expressed as follows:
\vspace{-5mm}
\begin{equation}
    \mathcal{L}_{\text{a-l}} = -\frac{1}{m} \sum_{v_i \in V} \log \left( \frac{\sum_{v_j \in \text{\posset}_{v_i}} e^{\text{sim}(\mathbf{z}_{v_i}^{a}, \mathbf{z}_{v_j}^{l})/\eta}}{\sum_{v_j \in (\text{\posset}_{v_i} \cup  \text{\negset}_{v_i})} e^{\text{sim}(\mathbf{z}_{v_i}^{a}, \mathbf{z}_{v_j}^{l})/\eta}} \right),
\end{equation}
where $\eta$ is a temperature parameter. Similarly, loss for contrasting the node representation from the global structure-driven view \(\mathbf{Z}^g\) with the local structure-driven view \(\mathbf{Z}^l\) can be expressed as:
\vspace{-5mm}
\begin{equation}
    \mathcal{L}_{\text{g-l}} = -\frac{1}{m} \sum_{v_i \in V} \log \left( \frac{\sum_{v_j \in \text{\posset}_{v_j}} e^{\text{sim}(\mathbf{z}_{v_i}^{g}, \mathbf{z}_{v_j}^{l})/\eta}}{\sum_{v_j \in (\text{\posset}_{v_i} \cup  \text{\negset}_{v_i})} e^{\text{sim}(\mathbf{z}_{v_i}^{g}, \mathbf{z}_{v_j}^{l})/\eta}} \right).
\end{equation}

Finally, loss for contrasting the attribute-driven view \(\mathbf{Z}^a\) and the global structure-driven view \(\mathbf{Z}^g\), is defined as below:
\begin{equation}
    \mathcal{L}_{\text{a-g}} = -\frac{1}{m} \sum_{v_i \in V} \log \left( \frac{\sum_{v_j \in \text{\posset}_{v_i}} e^{\text{sim}(\mathbf{z}_{v_i}^{a}, \mathbf{z}_{v_j}^{g})/\eta}}{\sum_{v_j \in (\text{\posset}_{v_i} \cup   \text{\negset}_{v_i})} e^{\text{sim}(\mathbf{z}_{v_i}^{a}, \mathbf{z}_{v_j}^{g})/\eta}} \right).
\end{equation}

Thus, the total contrastive loss \( \mathcal{L}_{\text{con}} \) can be expressed as  $\mathcal{L}_{\text{con}} = \mathcal{L}_{\text{a-l}} + \mathcal{L}_{\text{g-l}} + \mathcal{L}_{\text{a-g}}.$ Here, the negative samples are selected using either the distance-based ($\mathcal{N}_{dis}(v_i)$) or similarity-based ($\mathcal{N}_{sim}(v_i)$) sampling strategies. To train the model end-to-end, we combine the contrastive loss $\mathcal{L}_{\text{con}}$ with a supervised loss $\mathcal{L}_{\text{sup}}$, which is a standard cross-entropy loss. The final loss $\mathcal{L}$ can be expressed as: $\mathcal{L} = \mathcal{L}_{\text{con}} + \mathcal{L}_{\text{sup}}.$

\section{Experiment}
\subsection{Experimental Setup}
We evaluate \hypergcl\ on five diverse datasets, with statistics and hypergraphs detailed in Table \ref{dataset_statistics}. The datasets include Cora, Citeseer (CS), Wiki, Twitch-PT (PT), and LastFMAsia (LFMA), all sourced from PyTorch Geometric \cite{fey2019fast}. After an exhaustive search, global node counts are set at 3, 1, 4, 5, and 4. An overlapping community detection algorithm \cite{chen2010detecting} is used with default parameters. The data is split into: 10\% for training, 10\% for validation, and 80\% for testing.

\hypergcl\ is compared against sixteen baseline models, including graph-based models (GCN \cite{kipf2016semi}, GAT \cite{velivckovic2017graph}, GraphSage \cite{hamilton2017inductive}, DGI \cite{velivckovic2018deep}, GMI \cite{peng2020graph}, MVGRL \cite{hassani2020contrastive}, GraphCL \cite{You2020GraphCL}, GraphMAE \cite{hou2022graphmae}) and hypergraph-based models (HGNN \cite{feng2019hypergraph}, HCHA \cite{bai2021hypergraph}, HyperGCN \cite{yadati2019hypergcn}, DHGNN \cite{jiang2019dynamic}, HNHN \cite{dong2020hnhn}, UniGCNII \cite{huang2021unignn}, AllSetTransformer \cite{chien2021you}, DHKH \cite{kang2022dynamic}). Baseline hypergraphs are constructed following the original methodologies.  The baselines are considered if their experimental results or codes are available.

Local connectivity information ($lce$) is integrated using a two-layer GCN implemented in DGL \cite{wang2019dgl}. For computing $ce$ and $de$, we use learnable encoding functions based on PyTorch's \textit{Embedding layer} \cite{paszke2019pytorch}. A single-layer \hypergcl\ model is trained using Adam, with hyperparameters tuned via grid search on the validation set. Experiments are conducted with ten random splits, using one-hot encoded node and hyperedge initial features. Key hyperparameters include a learning rate of 0.001, dropout rate of 0.1, $k = 50$ (for $k$-NN) and $k = 60$ (for $k$-means), $s = 2$, $\tau = 0.2$, $\theta = 0.8$, $t = 25$, and $\eta = 0.5$. LeakyReLU activation, two attention heads, and early stopping after 100 epochs are applied. Both \hygan\ and \shygan\ use a hidden dimension of 64. All experiments are implemented in DGL with PyTorch and executed on an NVIDIA L40S-46GB GPU.

\subsection{Performance Comparison}
The results of our model, along with those of selected baselines, are presented in Table \ref{result}. These results demonstrate the consistent superiority of our model across all datasets.
Specifically, our model \hypergcld\ with distance-based negative samples, excels on the Cora dataset, achieving an impressive accuracy of 85.88\%. This significantly surpasses the accuracy of the best-performing graph-based baseline model, GraphMAE, at 83.80\% and exceeds the top-reported accuracy of the hypergraph-based baseline, DHGNN, which stands at 72.22\%. In the case of the Citeseer dataset, our model attains an accuracy of 73.12\%, outperforming the graph-based leading baseline GMI with an accuracy of 73.0\%, and the hypergraph-based top-performing baseline, AllSetTransformer, at 66.60\%. The trend continues with the Wiki, Twitch-PT, and LastFMAsia datasets, where our model substantially outperforms the baselines. The results underscore our model's substantial enhancements in classifying the datasets, setting a new standard compared to existing state-of-the-art methods. 

Moreover, this table shows that \hypergcl\ with distance-based negative samples \hypergcld\ performs better than similarity-based negative samples \hypergcls. Distance-based negative sampling chooses negative samples for a node based on network connectivity information, whereas similarity-based negative sampling uses node feature information to choose negative samples. Thus, based on the performance, we can infer that information on network connectivity is more important.


A closer look at Table \ref{result} reveals that hypergraph-based models generally lag behind the top-performing graph-based models. Traditional HyperGNNs are effective at capturing higher-order global structural information from the data. However, they might miss some important local structural information as they do not consider local connection details. Additionally, the baseline models typically create hypergraphs based on a single aspect of the underlying data. In contrast, our approach generates different types of hypergraphs by leveraging multiple aspects of the input data.
Nonetheless, hypergraph-based models like DHGNN, AllSetTransformer, and DHKH show better performance compared to other hypergraph-based models. Specifically, DHGNN and DHKH simultaneously learn the hypergraph structure and hypergraph neural network, enabling them to prune noisy and task-irrelevant connections, thus improving performance. The AllSetTransformer framework, which blends Deep Sets and Set Transformers with hypergraph neural networks, offers substantial modeling flexibility and expressive power, enhancing performance in various tasks.
\begin{table}
\centering
\caption{Impact of different components of \hypergcl\ on the model performance (accuracy \%).}
\normalsize
\begin{tabular}{|C{2.0cm}|C{0.75cm}|C{0.7cm}|C{0.8cm}|C{0.70cm}|C{0.9cm}|}
\hline
\cellcolor{gray!60}\textbf{\hypergcl \scriptsize{W/O}} &
\cellcolor{gray!60} \textbf{Cora} &
\cellcolor{gray!60} \textbf{CS} &
\cellcolor{gray!60} \textbf{Wiki} &
\cellcolor{gray!60} \textbf{PT} &
\cellcolor{gray!60} \textbf{LFMA} 
\\
\hline
$\mathcal{H}^{a}$  & 83.15 & 71.23 & 67.74 & 68.11 & 81.98\\
\hline
$\mathcal{H}^{l}$  & 83.78 & 71.05 & 67.92 & 67.25 & 83.11\\
\hline
$\mathcal{H}^{g}$ & 82.65 & 70.84 & 66.89 & 68.23 & 82.42\\
\hline
\hline
Augmentation & 83.88 & 72.36 & 67.52 & 67.78& 83.20\\
\hline
\netcl & 82.45 & 72.03 & 67.88 & 68.17 & 83.89\\
\hline
\shygan & 84.05 &72.28 & 68.49 & 68.66 & 84.29\\
\hline
\textbf{\hypergcld} & \textbf{85.88} & \textbf{73.12} & \textbf{69.22} & \textbf{70.10} & \textbf{85.15}\\
\hline 
\end{tabular}
\label{case1}\vspace{-5mm}
\end{table}

\subsection{Ablation Study}
To evaluate the contribution of different components in \hypergcl, we perform an ablation study as follows:

\textbf{i. Impact of Hypergraph Views} 
\hypergcl\ incorporates three distinct hypergraph views, each capturing unique aspects of the underlying graph. To assess the impact of each view, we remove each view in turn from \hypergcld\ and compare performance. Table~\ref{case1} shows that discarding the global structure-driven hypergraph view \(\mathcal{H}^g\) leads to the most significant performance drop, underscoring the importance of capturing global structural patterns for contrastive learning.

\textbf{ii. Impact of Adaptive View Augmentation.}
\hypergcl\ employs a learnable Gumbel-Softmax function to adaptively augment each hypergraph view, generating robust samples and selectively emphasizing critical relationships for contrastive learning. To quantify its effect, we remove it and compare the results against our main model across all datasets. As presented in Table~\ref{case1}, the absence of adaptive augmentation results in a noticeable drop in performance, highlighting its importance in producing diverse and informative training examples. 

\textbf{iii. Impact of \netcl}
Many existing GCL methods adopt a vision-inspired contrastive loss, treating an anchor node and its multiple views as positive samples, and all other nodes as negatives. This approach disregards the underlying network structure. In contrast, our proposed \netcl\ integrates connectivity information to more accurately define positive and negative samples. To assess its effectiveness, we remove \netcl\ and revert to the vision-inspired approach where each node's alternative views are positive samples and all others are negatives. As shown in Table~\ref{case1}, excluding \netcl\ degrades the model’s performance, underscoring the importance of incorporating structural cues into contrastive objectives.

\textbf{iv. Impact of \shygan\ and its components} We employ view-specific encoders for each hypergraph view: \hygan\ for \(\mathcal{H}^{a}\), and a specialized variant, \shygan, for \(\mathcal{H}^{l}\) and \(\mathcal{H}^{g}\). \shygan\ enhances node representations by incorporating learnable structure encodings and employs a topology-guided attention mechanism to identify important nodes and hyperedges from both semantic and structural perspectives.  
First, we evaluate the effect of \shygan\ by replacing it with \hygan\; the results in Table~\ref{case1} show a significant performance degradation. Additionally, to understand how each component of \shygan\ contributes, we remove them one at a time and test them on Cora and LastFMAsia. Table~\ref{shygan} indicates that excluding local structure encoding (\(\textit{lse}\)) leads to the largest performance drop, underscoring its vital role in preserving local connectivity lost when forming community-based hyperedges. This result demonstrates that capturing fine-grained local structure via a GCN is crucial for maintaining overall performance. 

\begin{figure}[t]
    \centering
\includegraphics[width=7.5cm, height=3.7cm]{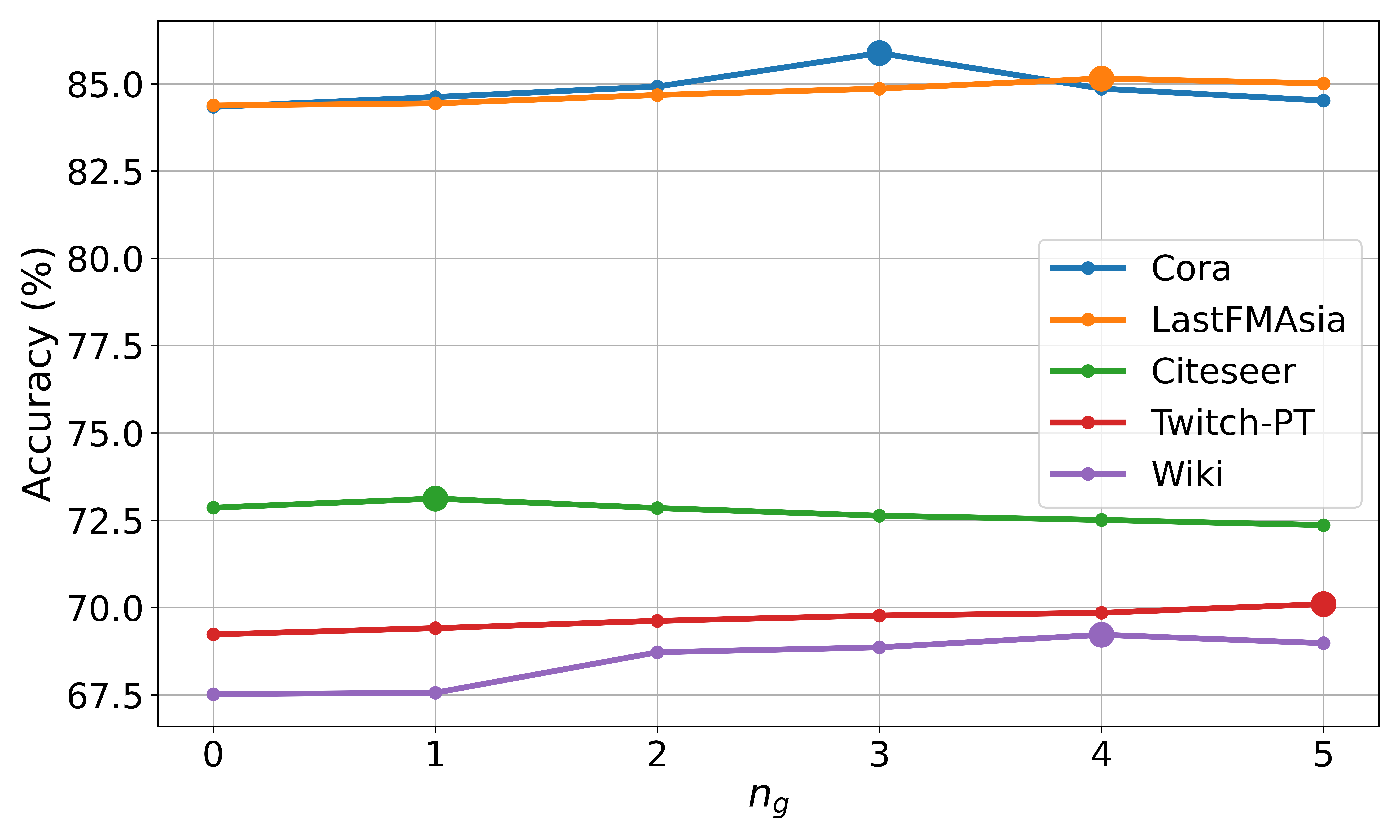}\vspace{-1mm}
   \caption{The performance (accuracy \%) of \hypergcl\ with different numbers of global nodes ($n_g$) in \(\mathcal{H}^g\).}
    \label{fig:line_graph}\vspace{-5mm}
\end{figure}

\begin{table}[t]
  \centering
  \caption{Impact of different components of \shygan\ on the model performance (accuracy \%).}\vspace{-2mm}
  \label{ablation2}

  \setlength{\tabcolsep}{2pt} 
  \renewcommand{\arraystretch}{1.2} 
  \begin{tabularx}{\linewidth}{|c|*{5}{>{\centering\arraybackslash}X|}}
    \hline
    \cellcolor{gray!60}\diagbox[dir=NW,innerwidth=1.8cm]{\textbf{Dataset}}{\textbf{\shygan} \\ \scriptsize{W/O}} & \cellcolor{gray!60}  \textbf{$lce$} & \cellcolor{gray!60} \textbf{$ce$} & \cellcolor{gray!60} \textbf{$de$} & \cellcolor{gray!60} \textbf{$lc$} & \cellcolor{gray!60}  \textbf{$hd$} \\
    \hline
    Cora & 84.19 & 84.73 & 85.15  & 84.45 & 84.98  \\
    \hline
    LFMA  & 84.36 & 84.68  & 85.01 & 84.86 & 84.66  \\
    \hline
  \end{tabularx} \label{shygan} \vspace{-1mm}
\end{table}
\textbf{v. Impact of global nodes} 
We investigate how the number of global nodes (\(n_g\)) in \(\mathcal{H}^g\) influences model performance, as illustrated in Figure~\ref{fig:line_graph}. For the Cora dataset, accuracy rises with \(n_g\), peaking at 85.88\% when \(n_g=3\) before declining. A similar pattern is observed for LastFMAsia, achieving its highest accuracy of 85.15\% at \(n_g=4\). For Citeseer, Wiki, and Twitch-PT, the optimal values of \(n_g\) are 1, 4, and 5, respectively. These results suggest that selecting an optimal number of global nodes is crucial for effectively incorporating global context without introducing excessive parameters that can degrade generalization. 
\section{Conclusion}
This paper introduces \hypergcl\, a novel Graph Contrastive Learning (GCL) framework that leverages three distinct hypergraph views to capture comprehensive attribute and structural information. By using a learnable Gumbel-Softmax function for adaptive augmentation and integrating a network-aware contrastive loss (\netcl), \hypergcl\ addresses critical limitations in existing GCL methods. Extensive experiments on benchmark datasets demonstrate that \hypergcl\ achieves state-of-the-art performance in node classification tasks, significantly outperforming both traditional graph-based and hypergraph-based models. The ablation studies confirm the critical role of each component in our framework, highlighting its robustness and adaptability across diverse datasets.
\bibliographystyle{IEEEtran}

\bibliography{main}
\end{document}